\documentclass[letterpaper, 10 pt, journal, twoside]{IEEEtran}
\IEEEoverridecommandlockouts                              

\RequirePackage[english]{babel}				
\RequirePackage[utf8]{inputenc}				
\RequirePackage[T1]{fontenc}				
\RequirePackage{amsmath,amssymb,amsfonts}	
\RequirePackage{graphicx}					
\RequirePackage{subfigure}					
\RequirePackage{float}						
\RequirePackage{siunitx}					
\RequirePackage{booktabs}					
\RequirePackage{tabularx}					
\RequirePackage{color}						
\usepackage{gensymb}
\usepackage{siunitx}
\usepackage{cite}							
\usepackage{multirow}
\usepackage{array}
\newcommand{\kj}[1]{\textcolor{black}{#1}}

\title{
Transport and Delivery of Objects with a Soft Everting Robot
}

\author{Ethan M. DeVries*, Jack Ferlazzo, Mustafa Ugur, Laura H. Blumenschein%
\thanks{Received 13 June 2025; accepted 29 December 2025. Date of publication 19 January 2026; date of current version 27 January 2026.  This paper was recommended for publication by Editor Yong-Lae Park upon evaluation of the Associate Editor and Reviewers' comments.}
\thanks{This work is supported in part by National Science Foundation grants 2308653.}%
\thanks{All authors are with the School of Mechanical Engineering, Purdue University, 47907, IN, USA. $^*$Corresponding author. 
        {\tt\small \{ edevrie, jferlazz, mugur, lhblumen\}@purdue.edu}}%
\thanks{Digital Object Identifier (DOI): see top of this page.}
}

\begin{document}

\maketitle

\markboth{IEEE ROBOTICS AND AUTOMATION LETTERS, VOL. 11, NO. 3, MARCH 2026} {DeVries \MakeLowercase{\textit{et al.}}: Transport and Delivery of Objects with a Soft Everting Robot}

\begin{abstract}

Soft everting robots present significant advantages over traditional rigid robots, including enhanced dexterity, improved environmental interaction, and safe navigation in unpredictable environments. While soft everting robots have been widely demonstrated for exploration type tasks, their potential to move and deliver payloads in such tasks has been less investigated, with previous work focusing on sensors and tools for the robot. Leveraging the navigation capabilities, and deployed body, of the soft everting robot to deliver payloads in hazardous areas, e.g. carrying a water bottle to a person stuck under debris, would represent a significant capability in many applications. In this work, we present an analysis of how soft everting robots can be used to deliver larger, heavier payloads through the inside of the robot. We analyze both what objects can be delivered and what terrain features they can be transported through. Applying existing models, we present methods to quantify the effects of payloads on robot growth and self-support, and we modified model to predict payload slippage. We then experimentally quantify payload transport using a soft everting robot with a variety of payload shapes, sizes, and weights and though a series of tasks: delivery, steering, vertical transport, movement through holes, and movement across gaps. We can transport payloads in various shapes and up to \SI{1.5}{\kilogram} in weight. We show that unsteered growth can deliver objects with high positional accuracy and precision in uncluttered environments and moderate accuracy in cluttered environments. We can move through circular apertures with as little as \SI{0.01}{\centi\meter} clearance around payloads. We can carry out discrete turns of up to \SI{135}{\degree} and move across unsupported gaps of up to \SI{1.15}{\meter} in length.
\end{abstract}


\vspace{-1.5 em}
\section{INTRODUCTION}\label{sec:intro}

\IEEEPARstart{T}{he} field of soft robotics has shown that compliance and novel locomotion modes can be used to navigate complex environments~\cite{rus2015design, laschi2016soft, della2020soft}. One versatile class of soft robots is the soft everting robot (i.e. Vine Robots~\cite{vineRobotsReview,blumenschein2020design}), which extends its body by everting a thin, flexible tube from its tip while remaining tethered to a base station.
Only the tip moves during growth, reducing friction and environmental disturbance~\cite{hawkes2017soft}. Soft everting robots have been shown to traverse highly constrained and cluttered spaces, making them promising for use in search-and-rescue operations \cite{roboa}, medical procedures \cite{giri2025inchigrab}, and exploration of underwater or underground environments \cite{coralReefs, SoftBurrowingRobot}. 
While many sensing and actuation strategies have been developed for soft everting robots \cite{vineRobotsReview, vineRobotsReview2}, significant opportunities remain, especially in leveraging the unique capabilities of the robot to carry payloads. Recent studies have addressed some interaction limitations, including stronger eversion drives for higher force output \cite{OseleICRA2024} and tip attachments for sensors and grippers \cite{suulker2023soft, kim2023soft, tipMount, softCameraHolder}. However, these works still use the soft everting robot tip to carry objects, limiting the carrying capacity and increasing the frictional interaction of the robot with the environment.

A similar class of robots, \kj{swallowing} grippers \cite{park2025soft, sui2022bioinspired, li2020bioinspired, li2025variable, root2021bio}, suggests a different way to deliver objects with these robots. \kj{Swallowing} grippers secure by everting and inverting a membrane around a target, generating holding forces through pressure, surface friction, and object geometry. If we imagine a soft everting robot as an extended \kj{swallowing} gripper \kj{moving in reverse}, payload delivery can be achieved by everting the internal membrane.

Early demonstrations of soft everting robots used the internal channel for small lightweight payloads, like small cameras, small circuit boards, and wires \cite{hawkes2017soft,steerableCamera}. This required extensive preplanning, as everything had to be preinserted into the system, limiting its versatility. For anything tethered, this also required access to the internal working channel to continually pull back on wires as the robot grew~\cite{blumenschein2020design}. As a result, this approach could not be used in designs which stored the material on a spool, thereby sealing off the one end of the internal working channel. An alternative base design, an origami‑inspired mechanism, introduced a way to compactly store material while maintaining the internal working channel for a camera and wire~\cite{origamiRyu}, and subsequent work redesigned the everting robot to open the channel and pass tools without friction \cite{Seo2024WorkingChannel}. These efforts have focused on tools extended through the channel, not analyzing the robot's ability to secure, transport, or deliver payloads, as recently shown~\cite{Boateng2024SSRR}. 

\begin{figure}[t]
\centerline{\includegraphics[width=0.8\columnwidth]{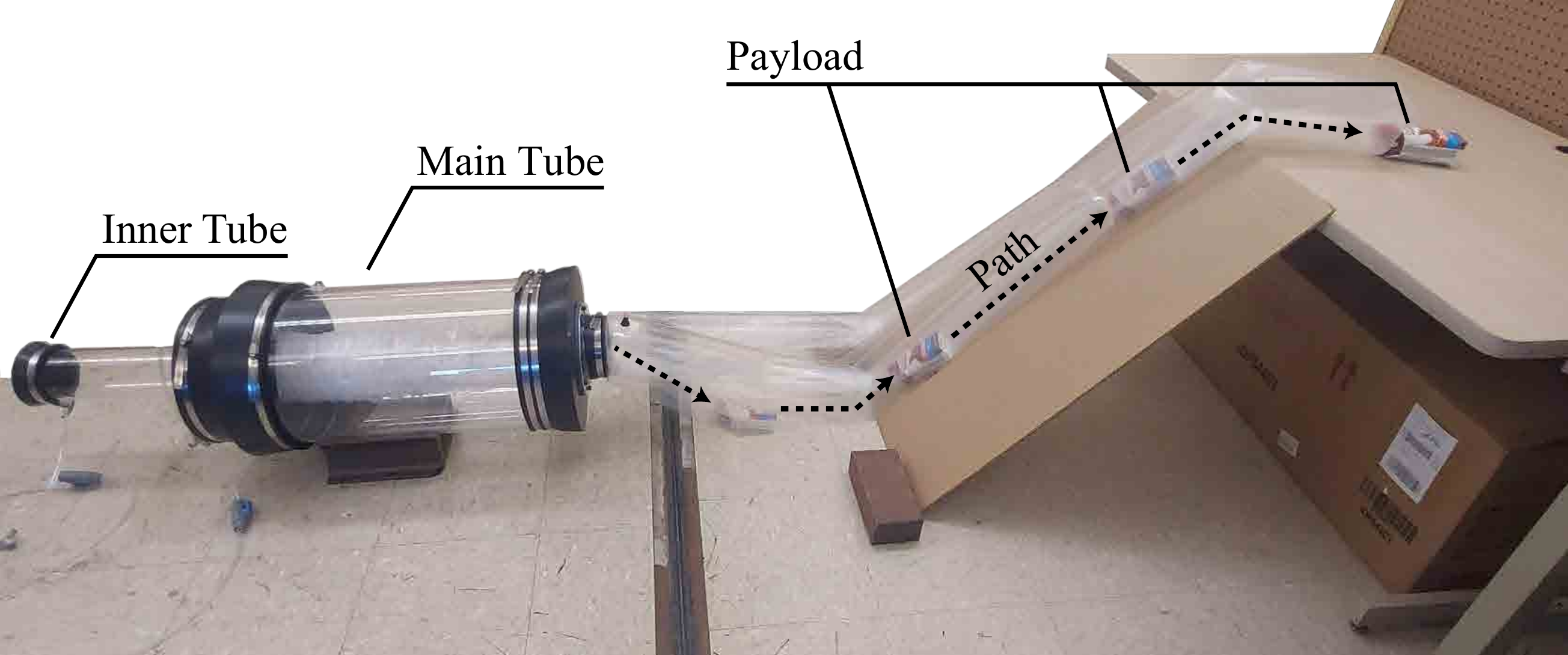}}
\vspace{-0.5 em}
\caption{Soft everting robot carrying an object up a $45$-degree slope.}
\label{fig-intro}
\vspace{-1.9 em}
\end{figure}

This paper addresses critical knowledge and design gaps in the use of internal object transport. We introduce a novel feature that enables continuous payload insertion into the robot’s internal working \kj{channel}, or tail as it grows, eliminating the need for preinserted or preplanned payloads (Fig. 1). Using existing models for soft everting robot motion and \kj{swallowing} grippers, we characterize payload delivery through the tail, of the soft everting robot, examining what features determine if the system can secure, transport, and deliver objects. The tubular body encapsulates a payload and it is moved forward toward the robot tip by eversion. We theoretically and experimentally show how various payload geometries and weights can be delivered and how payload delivery affects soft everting robot movement. The remainder of this paper is organized as follows. Section~\ref{sec:design} reviews the existing origami‑inspired soft everting robot designs \cite{origamiRyu} and discusses our adaptations for internal object transport and delivery.
Section~\ref{sec:modeling} then analyzes the effects of payloads on soft everting robot, using existing everting robot and gripper analytical models, in order to predict successful delivery. We experimentally validate these theoretical analyses and demonstrate payload transportation in Section~\ref{sec:experimental} and discuss results in Section~\ref{sec:discussion}.

\section{Design Overview} \label{sec:design}

In this section, we review the design concept for a soft everting robot with an accessible internal working \kj{channel}, first demonstrated by Kim \textit{et al.}~\cite{origamiRyu}, and discuss the materials and modifications made to deliver a variety of payloads.

\begin{figure}[t]
    \centering
    \includegraphics[width=0.85\columnwidth]{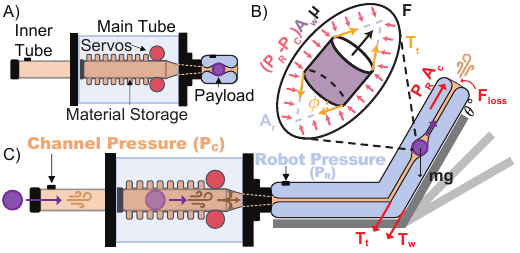}
    \caption{Payload-capable soft everting robot.  
    A) Labels for the key features and components of the system. B) A free-body diagram of the forces acting on the payload within the pressurized inner pressure zone. C) Illustration of the system with the \kj{internal working  channel} pressurized to \kj{$P_C$} (orange) and the surrounding robot pressurized to \kj{$P_R$} (blue). Shown are the forces acting during growth while carrying an internal payload. }
    \label{fig-design}
    \vspace{-1.6em}
\end{figure}

The Kim \textit{et al.}~\cite{origamiRyu} paper used an origami-inspired mechanism featuring an origami folding pattern to replace the previously used spool. We take inspiration from this but found that as we increased the inner tube diameter for payload insertion, self-organized folds wrapped around the rigid inner tube were more effective for storing the material(Fig.~\ref{fig-design}). Since the far end of material is attached to the inner tube, the internal working \kj{channel} remains open, allowing objects equal to or smaller than the inner tube to pass from the base to the tip throughout delivery. Servos are used to supply and retract the membrane from the tube. The internal working \kj{channel} applies a pressure envelope around any objects inside, pulling them along as the robot everts. This pressure is also applied on the inner tube, increasing friction, but a flow of pressurized air can be applied to the inner tube to helps with breaking the friction on the stored material.

To deliver individual payloads, we need to increase the inner tube’s size and make the outside end unsealable, allowing payloads to be added during growth. We aim to deliver payloads around the size of a water bottle, i.e. objects around \SI{6}{\centi\meter} in diameter, so we use an inner tube diameter, \(D_t\), of \SI{10.2}{\centi\meter}. We must also increase the diameter of the robot, \(D_{\text{robot}}\), to accommodate, increasing to \SI{12.7}{\centi\meter}. Comparing this to the earlier work by Kim \textit{et al}.~\cite{origamiRyu}, which used \(D_{\text{robot}} = \SI{5}{\centi\meter}\) and \(D_t = \SI{1.3}{\centi\meter}\), yields a ratio of \(D_{\text{robot}}/D_t = 3.8\), whereas our design achieves \(D_{\text{robot}}/D_t \approx 1.27\)

While this lower ratio lowers the packing efficiency of our material, increasing \(D_{\text{robot}}\) to preserve the larger aspect ratio also introduces drawbacks. Larger diameters reduce maneuverability by increasing the turning radius and limiting the robot’s fit through smaller apertures. They also become flow-rate limited~\cite{heap2024large}, requiring more air volume for growth. The burst pressure also decreases, limiting the pressure to secure and lift payloads (Section~\ref{sec:modeling}). Reducing the aspect ratio too close to 1:1 causes excessive friction \cite{godaba2019payload} and insufficient space for self-stacking, increasing resistance to material feeding and limiting stored length. We found that our chosen \(D_{\text{robot}}/D_t \approx 1.27\) balances these factors, accommodating meaningful payloads while maintaining efficient growth, manageable friction, and good mobility in confined spaces.

The remainder of the materials to construct the soft everting robot are as follows. The inner tube and main tube are made of polymethyl methacrylate (PMMA) (i.e. acrylic) tubes and all end caps are modified QwikCaps. The soft everting robot is made from a \SI{0.0508}{\milli\meter} thick Low-Density PolyEthylene (LDPE) tube. 
A low‑friction ultra‑high‑molecular‑weight polyethylene (UHMW) material lined the inner tube to lower friction. Two high‑torque continuous‑rotation servos (DYNAMIXEL XM540‑W270‑T, stall torque \SI{10.6}{\newton\meter} at \SI{12}{\volt}, no‑load speed \SI{30}{\text{rpm}}) were installed to pinch the membrane between silicone rollers and the tube and thereby overcome the remaining friction. The pressures in both inner and main tubes were controlled by QB3 pressure controllers and the robot pressure, $\kj{P_R}$ was measured by Honeywell Sensing 5 psi (SSCDANN005PGAA5) pressure sensor. Due to the open tip, we report the commanded \kj{working channel} pressure for $\kj{P_C}$.

\section{Theoretical Analysis}\label{sec:modeling}

This section applies existing quasi-static models from prior literature to theoretically analyze how a soft everting robot secures, transports, and delivers internal payloads. These models predict three key behaviors: the robot’s ability to grow, the payload’s security without slipping, and the robot’s stability without buckling. Each subsection draws from analytical frameworks for soft everting robots and related gripping systems and we assess predicted delivery performance and identify where additional physical effects, like payload geometry or slope angle, may affect accuracy.


\subsection{Object Transportation}\label{subsec:transport-model}

A soft everting robot’s growth starts when the tip pressure exceeds the resisting tension forces~\cite{coad2020retraction}. Heap \textit{et~al.} later expanded the model to include friction losses and flow-rate limitations~\cite{heap2024large}. We adapt their formulation to account for a payload of mass \(m_{\text{payload}}\) moving vertically along an incline of angle~\(\theta\):
\begin{equation}
    \kj{P_R} A_c = 2T_t + F_{\text{loss}} + (m_{\text{payload}} + m_{\text{robot}}) g \sin\theta .
    \label{eq:inclined_growth}
\end{equation}
where \(A_c = \tfrac{\pi}{4} D_{\text{robot}}^{2}\) is the tip cross sectional area, \(T_t\) is the tail tension, \(F_{\text{loss}}\) is the frictional loss at the tip during eversion, \(g\) is the acceleration due to gravity, \(\theta\) is the slope angle, \(m_{\text{robot}}\) is the mass of the robot, \(m_{\text{payload}}\) is the mass of the payload, and \(\kj{P_R}\) is the measured robot pressure. When the control system specifies a set pressure, pressure over the minimum value will be lost to friction proportional to air flow rate and to growth velocity~\cite{hawkes2017soft}. The maximum robot speed, \(v_{robot}\), will derive from the maximum \kj{channel} flow rate \(\dot{Q}_{\text{\kj{channel}}}\)\cite{heap2024large}.
Servo actuation was used to limit additional friction from the base and regulate material supply. If the servo feed speed exceeds twice the robot velocity (\(v_{\text{servo}}>2v_{\text{robot}}\)), the tail remains slack (\(T_t \approx 0\)); otherwise, pressure equalizes with the set point, maintaining steady growth.


\subsection{Object Securing}\label{subsec:secure-model}

To determine when objects can be held without slipping, the robot is modeled as an \kj{swallowing} gripper~\cite{sui2022bioinspired,li2020bioinspired,li2025variable,root2021bio,park2025soft}. Building on the securing model from Sui \textit{et~al.}~\cite{sui2022bioinspired}, the maximum load before slip arises from Coulomb friction and wall tension due to the wrap angle of the membrane. In our study, this pull-out force is reframed as the payload weight that can be transported without slip. The effective grip contact area $K$ represents the change in securing force per unit pressure, defined as $K = \frac{\Delta F}{\Delta P}$ [N\,kPa$^{-1}$]. The effective contact area will be~\cite{sui2022bioinspired}:
\begin{equation}
    K = \num{1e3} \,\mu\!\left(A_w + \tfrac12 A_c\cos\phi\right), \label{eq:K_model}
\end{equation}
where $A_w$ is the wetted contact area, \(\mu\) is the friction coefficient, and $\phi$ is the average wrap angle. The wrap angle can be found from considering the force balance radially. The right side of the equation represents wall tension transitioning from the payload edge to equilibrium. The left side represents pressure balance affecting the material during this transition, as seen in Fig.~\ref{fig-design} (B). 
\begin{equation}
T_{\mathrm{t}}\cos\phi=(\kj{ P_R}-\kj{ P_C})\;A_{r}\cos\phi
    \label{eq:Transition-Balance}
\end{equation}
with $A_{r}$ being the surface area of the plastic as it goes from touching the object to being at equilibrium, this equation balances inward pressure with outward material tension (Fig.~\ref{fig-design} (B)). We assume the material forms a cone, so $A_r=\pi D_{\mathrm{payload}}^2/4\cos\phi$, where $D_{\mathrm{payload}}$ is the payload diameter. This \(T_t\) is the same as in Eqn. (\ref{eq:inclined_growth}), but without  \kj{\(m_{\text{payload}}\)}. For stationary robot tests, $F_{\mathrm{loss}}$ is zero, so $T_{\mathrm{t}}=\tfrac{1}{2}\kj{P_R} A_{\mathrm{c}}$. Plugging this into Eqn. (\ref{eq:Transition-Balance}) and solving for the \kj{wrap angle} $\phi$ yields
\begin{equation}
    \phi\approx\arccos\!\left(\frac{2 D_{\mathrm{payload}}^{2}}{D_{\mathrm{robot}}^{2}}\left(1-\frac{\kj{ P_C}}{\kj{ P_R}}\right)\right).
    \label{eq:Phi}
\end{equation}
The last unknown for Eqn. \eqref{eq:K_model} is the $A_{w}$, which includes only where the material touches and therefore depends on the object geometry (see Fig.~\ref{fig-secure}). 


\subsection{Buckling over Unsupported Gaps}\label{subsec:terrain-model}
\label{subsubsec:gap-crossing}
Soft everting robots can self-support but will buckle under excessive weight or length. Prior analyses of inflated beams have defined the critical bending moment as a function of diameter and pressure~\cite{comer1963deflections,leonard1960structural} which has shown good prediction of soft everting robots without \(T_t\) \cite{mcfarland2023collapse, haggerty2019characterizing}. A recent formulation by McFarland \textit{et~al.}~\cite{McFarland2025gap} extends this to include \(T_t\) from a centered tail, with \(M_{\text{collapse}}\) as:
\begin{equation}
  M_{\text{collapse}} = \frac{1}{2}\frac{\kj{P_R} \pi D_{\text{robot}}^{3}}{8}+\frac{F_{loss}D_{\text{robot}}}{4}.
  \label{eq:critical-collapse-moment}
\end{equation}
Collapse will occur from weight-induced bending moment, where the robot mass is a function of length~\cite{McFarland2025gap} where $\rho$ is the material density and $t$ is the membrane thickness:
\begin{equation}
  m_{\text{robot}} = 2 \pi D_{\text{robot}} t L_{\text{robot}} \rho .
  \label{eq:robot-mass}
\end{equation}
\begin{figure}[t]
\centerline{\includegraphics[width=0.85\columnwidth]{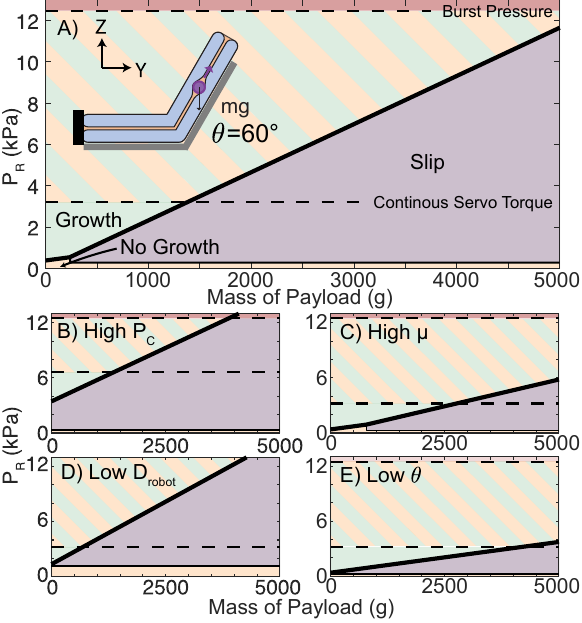}}
\vspace{-1em}
\caption{Behavior zones during slope climbing for a \SI{4.35}{\centi\meter} PETG cube payload on a \SI{60}{\degree} incline \kj{(shown by inset)}. A) shows the base case with no channel pressure; orange indicates \kj{$P_R$ is too low for growth to occur}, green - growth with secure payload, purple, payload slip, orange–green stripes, the servo torque limit, and red, burst pressure. B–E) show effects of \kj{B)} increasing \kj{$P_C$}, \kj{C) increasing} $\mu$, \kj{D)} decreasing $D_{robot}$, \kj{or E)} decreasing $\theta$.}
\label{fig-Incline-Analysis}
\vspace{-1.0 em}
\end{figure}
\begin{figure}[t]
\centerline{\includegraphics[width=0.85\columnwidth]{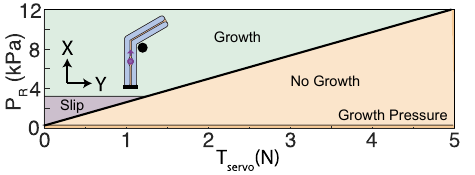}}
\vspace{-1em}
\caption{Behavior zones during payload transport through a \kj{$90^\circ$} bend \kj{around an obstacle} \kj{(shown by inset)}. \kj{A mix of robot pressure and servo torque is needed to achieve successful movement Values are based on} a \SI{6}{\centi\meter} PETG sphere at \kj{a constant} \SI{1}{\kilo\pascal} \kj{channel pressure}. Orange indicates no growth, green successful growth around the bend, and purple payload slip \kj{from contact with the obstacle}.}
\label{fig-Bend-Analysis}
\vspace{-1.0 em}
\end{figure}
The payload mass, $m_{payload}$, will add a moment based on its distance along the length, $d_{payload}$ and will also cause the tail to be offset to a diameter $D_{tail}$. Summing moments about the base and adding these modifications gives
\begin{multline}
  \frac{\kj{P_R} \pi D_{\text{robot}}^{2}(D_{robot}-D_{tail})}{8}+\frac{F_{loss}D_{\text{tail}}}{2}
  = \\ m_{\text{robot}} g \frac{L_{\text{robot}}}{2}
    + m_{\text{payload}} g d_{\text{payload}} .
  \label{eq:sum-moments}
\end{multline}
\noindent
This can be solved for a critical length which depends on pressure, payload mass, and payload location.

\subsection{Combined Model Predictions}\label{subsec:combined-models}

To visualize the interaction between derived models, we combine pressure, friction, servo, and geometric constraints to predict system behavior. Figure~\ref{fig-Incline-Analysis} shows modeled outcomes for slope climbing based on Eqns.~(\ref{eq:inclined_growth}) and (\ref{eq:K_model})-(\ref{eq:Phi}). This reveals areas of stable growth and payload slippage, all limited by servo torque and burst pressure. Figure~\ref{fig-Bend-Analysis} further predicts regimes for payload transport around a bend, using Eqns.~(\ref{eq:inclined_growth})-(\ref{eq:Phi}) and calculated as a function of servo tension and robot pressure. We predict that contact by the environment on the payload can cause stall, or slip. These figures identify feasible operating zones for payload transport in soft everting robots.

\section{EXPERIMENTAL}\label{sec:experimental}

In this section, we perform a series of experiments to validate the above analysis and to show the range of object delivery tasks that a soft everting robot can carry out.
\subsection{Object Transportation}
\label{subsec:transport-experiment}

\subsubsection{Object delivery}
\label{subsubsection:Delivery}
\begin{figure}[t]
\centerline{\includegraphics[width=.75\columnwidth]{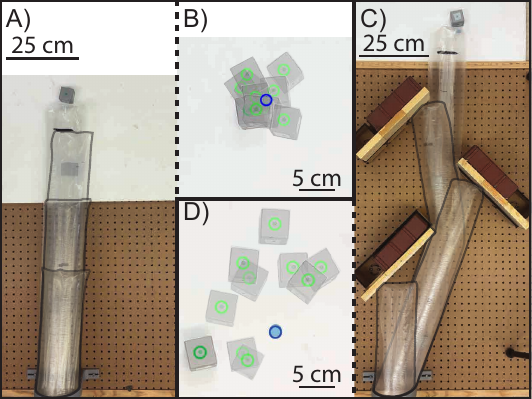}}
\caption{Payload delivery performance and accuracy metrics. A) Uncluttered environment. B) Delivery results with MRE = \SI{3.52}{\centi\meter}, RSD = \SI{3.61}{\centi\meter} \kj{from the uncluttered environment}. C) Cluttered environment. D) Delivery results with MRE = \SI{9.13}{\centi\meter}, RSD = \SI{7.81}{\centi\meter} \kj{from the cluttered environment}.
}
\label{fig-delivery}
\vspace{-1.0 em}
\end{figure}
To evaluate how accurately the robot can deliver a payload to a specified location, we conducted ten trials delivering a \SI{4.35}{\centi\meter} payload in both uncluttered and cluttered environments. In the first experiment (Fig.~\ref{fig-delivery}(A)) the robot delivered the payload across open ground to a target point. The payload was inserted when a black line passed the servos. The results (Fig~\ref{fig-delivery}(B)) show a Mean Relative Error (MRE) of \SI{3.52}{\centi\meter} and a Relative Standard Deviation (RSD) of \SI{3.61}{\centi\meter}. Considering a \kj{\(D_{robot}\)} of \SI{12.7}{\centi\meter}, these results demonstrate consistent and repeatable delivery performance within a small fraction of the robot’s body width.

The uncluttered condition served as a baseline for a second experiment conducted in a cluttered environment (Fig.~\ref{fig-delivery}(C–D)). The presence of obstacles introduced interactions between the robot, payload, and environment, such as changes in local buckling position, variable capstan friction around corners, and payload slippage after contact (as predicted in Fig.(\ref{fig-Bend-Analysis}), all of which affected delivery accuracy. Despite these additional challenges, the system achieved an MRE of \SI{9.13}{\centi\meter} and an RSD of \SI{7.81}{\centi\meter}, keeping the delivered payload within one robot diameter of the target in all cases. These results suggest that even when navigating cluttered terrain, the soft everting robot can deliver objects with moderate precision without active control. While the robot tip reliably reached the same spot, variations arose primarily from release-induced bouncing behavior due to changes in payload position and alignment in the tail after environment contact.

\subsubsection{Object transport on inclined surfaces}
\label{subsubsection:Incline-Surface}

\begin{figure}[t]
\centerline{\includegraphics[width=.85\columnwidth]{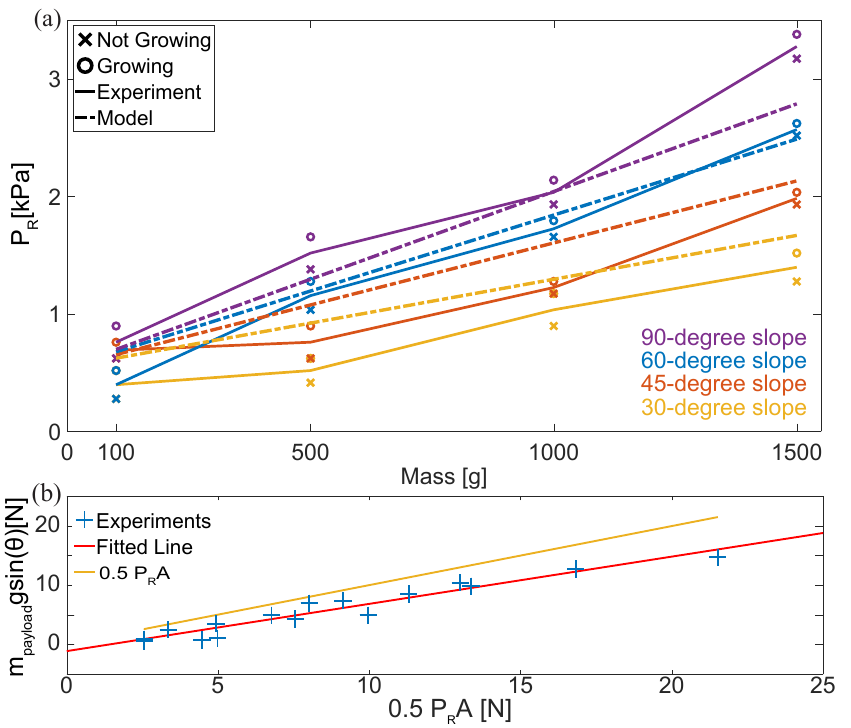}}
\caption{A) Maximum payload mass transported up an incline as a function of \(\kj{ P_R}\).  Individual curves represent slopes of \(30^{\circ}\), \(45^{\circ}\), \(60^{\circ}\), and \(90^{\circ}\) and model is plotted for comparison. B) Corresponding values of \(0.5\,\kj{ P_R} A_c\) plotted against \(m_{\text{payload}} g \sin\theta\) for all collected data. Best fit intercept predicts a \(F_{loss}\) value of \SI{3.6}{\newton}.
}
\label{fig-different-slopes}
\vspace{-1.0 em}
\end{figure}

We verify the effects of \kj{\(m_{payload}\)} on vertical growth, as modeled in Eqn.~(\ref{eq:inclined_growth}). A soft everting robot lifted a spherical payload on an angled slope. The payload’s mass varied from \SIrange{100}{1500}{\gram}, and the slope angle varied from \SIrange{30}{90}{\degree}. A $\kj{ P_C}$ value of \SI{1.4}{\kilo\pascal} reduced servo strain. With the payload on the slope, $\kj{ P_R}$ was raised until growth started, then lowered until motion stopped, giving lower and upper bounds on the growth threshold. 

Figure~\ref{fig-different-slopes}(a) shows thresholds and model lines.  Larger mass and steeper slope require higher $\kj{ P_R}$. Figure~\ref{fig-different-slopes}(b) plots $0.5 \kj{ P_R} A_{\mathrm{c}}$ against $m_{\text{payload}} g \sin\theta$ for all data, yielding a linear fit with vertical offset of \SI{3.6}{\newton}. This offset matches the predicted \(F_{loss}\) using equivalent material thickness (\SI{4.1}{\newton}~\cite{blumenschein2019design}), though the slight increase in offset at higher pressures is consistent with the predicted increase to prevent payload slip. Using this \(F_{loss}\), the minimum eversion pressure without a payload is \SI{0.284}{\kilo\pascal}. The model significantly underestimated \(\kj{ P_R}\) in pure vertical growth due to higher capstan friction. Payloads heavier than \SI{500}{\gram} contacted the corner and stalled, so for the \SI{90}{\degree} slope, the payload was placed above the fold. 


\subsubsection{Object transport through turns}
\label{subsubsection:turning-Surface}

\begin{figure}[t]
\centerline{\includegraphics[width=.85\columnwidth]{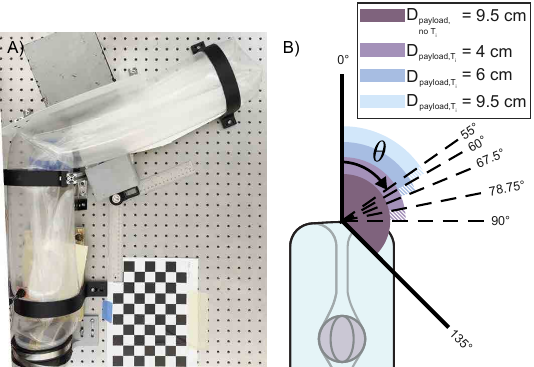}}
\caption{A) The test rig showing a (blue) payload that failed to traverse a fixed angle bend when tail tension was present. B) Experimental results for bend angles from \(45^{\circ}\) to \(135^{\circ}\). With \kj{\(T_t\)}, larger \kj{\(D_payload\)} could only pass acute angle turns (\(<90^{\circ}\)).  With \kj{\(T_t\)} removed every payload traversed all tested angles without jamming.}
\label{fig-bends}
\vspace{-1.15 em}
\end{figure}

A soft everting robot steers by buckling material on the inner surface, curving the body toward the buckled side. This introduces wrinkles that distort the internal cross-section, hindering payload movement.  While most steering is \emph{continuous}, produced by small distributed actuation \emph{discrete}  steering concentrates the bend at a single location, resulting in larger internal wrinkles. We focus on payload delivery around discrete bends as a worst-case scenario for potential contact on the payload and stall or slip.

Payloads of varying length and diameter were grown around increasing bend angles (\(\theta\)) until they could no longer be grown (Fig.~\ref{fig-bends}(A)). Tests were performed with and without \(T_t\).Walls constrained the robot’s bend angle, and a rigid bracket taped to the tip prevented backward motion, mimicking a mid-delivery bend. The final successful and first failed angles were noted. Performance didn’t change with test lengths, so results show successful angle range as a function of \kj{\(D_{payload}\)} (Fig.~\ref{fig-bends}(B)).  Larger \kj{\(D_{payload}\)} consistently contacted and jammed at smaller angles with \(T_t\)(i.e. higher \(T_{servo}\)). Eliminating \(T_t\) improved performance, and all tested lengths and diameters passed through the maximum bend angle, \(\theta = 135^{\circ}\). High \(T_t\) constricts the passage, causing interaction between payload and the wrinkled surface, as predicted in Fig.~\ref{fig-Bend-Analysis}.


\subsection{Object Securing}
\label{subsec: secure-experiment}

\begin{figure}[t]
\centerline{\includegraphics[width=0.75\columnwidth]{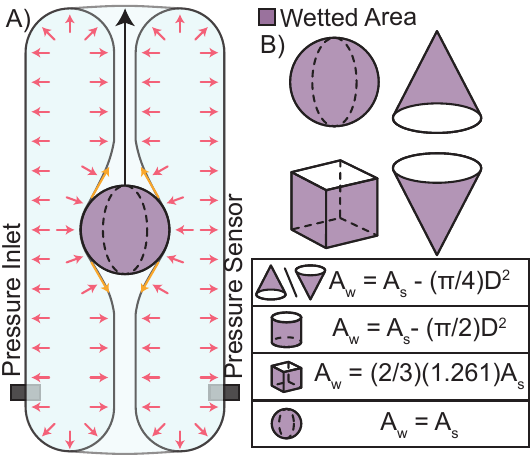}}
\caption{A) Experimental setup for the pull-out force measurement. B) Geometries used in the shape-series test, with \kj{\(A_w\)} highlighted in purple with a table of \kj{\(A_w\)} calculations by shape.}
\label{fig-secure}
\vspace{-1.0 em}
\end{figure}
\subsubsection{Pull-out Force}

To initially validate the securing model in Eqn.(\ref{eq:K_model}), a short length of soft everting robot structure was secured to prevent growth (Fig.~\ref{fig-secure}). This isolated the gripping conditions dependent on $\kj{P_R}$ and membrane mechanical properties. Using this simplified geometry, we estimated normal forces and friction coefficients without growth dynamics. For each test, the gripper was pressurized to \SI{5}{\kilo\pascal} and conformed around a payload suspended on a fishing line. A motorized force test stand with a force transducer (M7-100, Mark-10) pulled the fishing line upward, forcing the payload to slip on the inside surface of the tail material. During each pull-out trial, the force stand advanced at \SI{50}{\milli\metre\per\minute} and stopped at full extension or a maximum force of \SI{50}{\newton}. $\kj{P_R}$ was first set to \SI{5}{\kilo\pascal} and slowly decreased from small inherent leaks. Each run provided a continuous record of the pull-out force ($F$) versus the \kj{swallowing} gripper pressure from the pressure sensor ($\kj{P_R}$). Two experimental series were executed:

\begin{figure}[t]
\centerline{\includegraphics[width=.9\columnwidth]{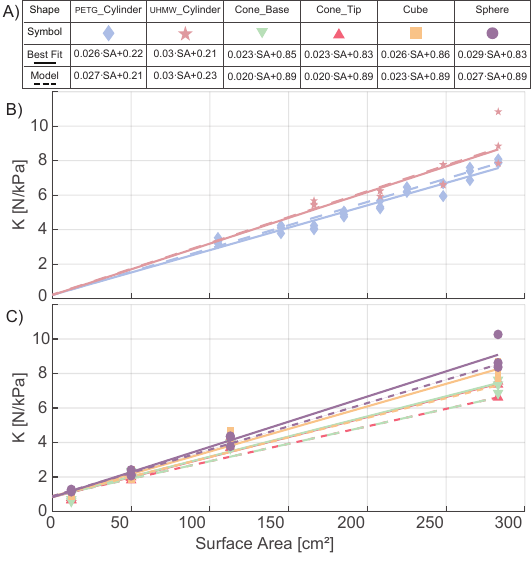}}
\caption{A) Legend summarizing all pull-out test groups, showing their symbols, experimental best-fit lines, and the theoretical curve from Eqn.~\eqref{eq:K_model}. B) Effective contact area ($ K \;=\; \Delta F/\Delta P$) during pull-out versus surface area for cylinders with different friction coefficients. C) Effective contact area during pull-out versus surface area for the various object geometries tested.}
\label{fig-secure-results}
\vspace{-1 em}
\end{figure}

\textbf{Surface Area series.} Cylinders of \SI{4.0}{\centi\metre} in diameter and lengths from \SIrange{6.0}{20.0}{\centi\metre} were printed from Polyethylene Terephthalate Glycol (PETG). Some were wrapped in UHMW tape to create two friction levels. Static friction versus LDPE was measured at $\mu_{\mathrm{PETG}}=0.27$ and $\mu_{\mathrm{UHMW}}=0.30$.

\textbf{Shape series.} Four geometries were tested: a sphere, a cube, and a cone tested base first and tip first. Four sizes of each shape were used with equal total surface area, based on sphere diameters of \SIlist{2.0;4.0;6.0;9.5}{\centi\meter}. 

The pull-out force data was linear with \kj{\(P_R\)}, so a least squares line was fit and the slope \(K\) extracted. The resulting \(K\) values are plotted (Fig.~\ref{fig-secure-results}(B) \& (C)). The surface area series (Fig.~\ref{fig-secure-results}(B)) show linear trends with slightly different slopes for PETG and UHMW. This matches Eqn. (\ref{eq:K_model}) and Fig. \ref{fig-Incline-Analysis}(C), as the relationship between \(A_w\) and \(K\) is governed by \(\mu\), so a larger \(\mu\) yields a steeper line. The intercept is set by the wrap component, which remains constant. Without the wrap component, the surface area alone underestimates \(K\) for short cylinders, where wrap becomes significant. Adding the wrap term brings the prediction within a mean absolute percentage error of \SI{6}{\percent} of PETG data and \SI{8}{\percent} of UHMW data. 

Fig.~\ref{fig-secure-results}(C) summarizes the shape series. Shape changes alter the \kj{\(A_w\)}, changing the slope as the membrane grips the surface differently. The \kj{\(A_w\)}s used in the model are shown in Fig.~\ref{fig-secure}. Rotationally uniform shapes have side projected surface \(A_w\), but cube’s sharp edges increase effective contact by a factor of \(1.261\), equivalent to the ratio of the square’s hydraulic-equivalent diameter to the nominal diameter.

\subsubsection{Payload Slippage}
\label{subsubsec:slippage}

With the model verified without \kj{\(P_C\)}, we now experimentally verify the conditions when payload slip begins inside the soft everting robot. Payload slippage will happen in the \kj{internal working channel pressure zone}. While it can limit transport (Fig.~\ref{fig-Incline-Analysis}), controlled payload slippage can delay delivery. We can decrease the securing force from pressure and use the environment or payload weight to reposition it. To prevent slippage, we increase the force needed. In this experiment, the servos were held in place to restrict growth. Any observable motion indicated payload slip, not growth. The robot with a constant $\kj{P_R}~(\SI{2.90}{\kilo\pascal})$ was placed on a $60^{\circ}$ incline while a spherical payload with mass \SIrange[]{200}{1000}{\gram} rested midway. The commanded $\kj{P_C}$ was increased stepwise, pausing to re-equilibrate and check for slip. When airflow caused slippage, $\kj{P_C}$ was lowered to stabilize the payload before continuing. The slip threshold for each mass was the lowest $\kj{P_C}$ that produced noticeable translation.

\begin{figure}[t]
\centerline{\includegraphics[width=.85\columnwidth]{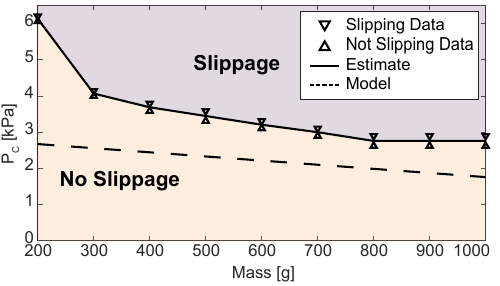}}
\caption{Plot showing \(m_{\text{payload}}\) versus \(\kj{P_C}\) for a \(60^{\circ}\) slope at a constant \(\kj{P_R} = \SI{2.90}{\kilo\pascal}\). The orange region marks pressure levels where the payload remains stationary, and the purple region marks pressure levels where it slips.}
\label{fig-slip-results}
\vspace{-1.5 em}
\end{figure}

Figure~\ref{fig-slip-results} plots \(m_{\text{payload}}\) against the slip threshold \(\kj{P_C}\). Triangles mark the upper and lower bounds of slippage, while the solid line marks the midpoint for an empirical slip boundary. The critical \(\kj{P_C}\) decreases monotonically with payload mass because heavier objects require less differential pressure to overcome static friction, which the model line matches well, though it under-predicts the necessary \(\kj{P_C}\), likely due to the air flow rate of the \kj{channel} pressure. The viable control window for \(\kj{P_C}\) narrows with increasing payload, but all payloads (Fig.~\ref{fig-different-slopes}) could be delivered with a \(\kj{P_C}\) of \SI{1.4}{\kilo\pascal}, so there remains a significant range of control across the full mass range tested. We tested how distributing the same mass across multiple payloads affected slip behavior. Three configurations totaling \SI{1.5}{\kilogram} were tested with one, two, or three spherical payloads in contact or separated by sealed sections. Dividing the mass had minimal effect on growth but significantly affected the robots’ ability to securely hold the payloads when not spaced apart. Incorporating a one‑dimensional flow analysis for \(\kj{P_C}\) into Eqn.~\eqref{eq:K_model} would yield a more accurate operating envelope for payload delivery.

\subsection{Challenging Terrain}
\label{subsec:terrain-experiment}


\subsubsection{Hole Restriction}
\label{subsubsec:hole-restriction-experiment}

\begin{figure*}[t]
\centerline{\includegraphics[width=1.8\columnwidth]{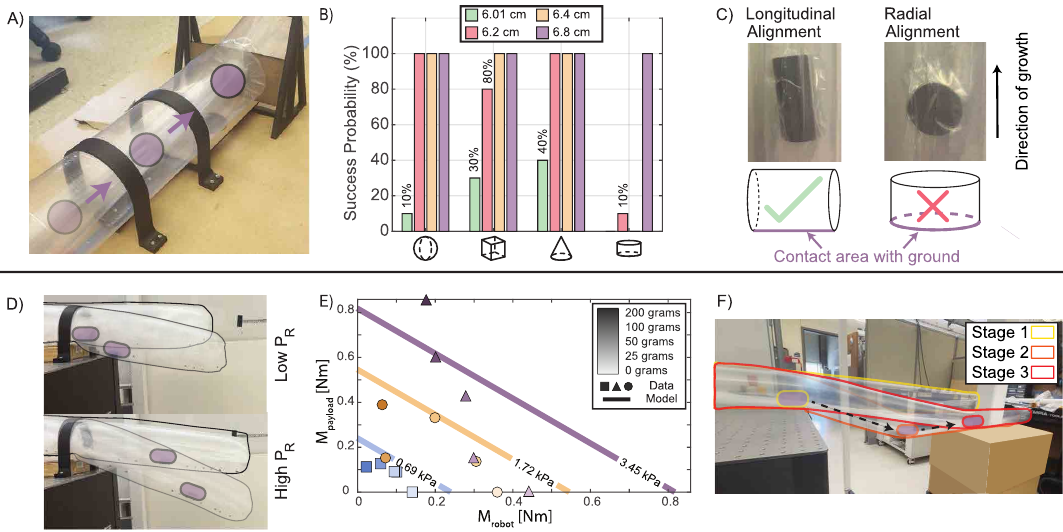}}
\caption{\kj{A-C}) Hole restriction experiments: 
\kj{A}) experimental setup showing the robot delivering a payload through a circular aperture,  
\kj{B}) success rate over 10 trials for each nominal \SI{6.0}{\centi\meter} shape (the cube is inscribed) passing through apertures of different diameters at a command \kj{robot} pressure of \SI{1.2}{\kilo\pascal},  
\kj{C}) images illustrating how payload geometry influences realignment inside the robot.  
\kj{D-F}) Gap crossing experiments:  
\kj{D}) images showing \kj{how increasing} \(T_t\), and \kj{the resulting increase of $P_R$}, \kj{improves} gap-crossing performance,  
\kj{E}) cantilever results with raw data and the model of Eqn.~(\ref{eq:sum-moments}), 
\kj{F}) demonstration that growing across the gap first, then transporting the payload, allows the robot to span larger gaps with heavier payloads.}
\label{fig-terrain-experiment}
\vspace{-1.0 em}
\end{figure*}

Soft everting robots can compress through apertures smaller than their body diameter, making them suitable for navigating confined spaces. To evaluate payload delivery in such environments, we quantified the robot’s ability to move objects through nearly size-matched circular apertures. We tested four objects with nominal diameters of \SI{6.0}{\centi\meter} against four hole diameters: \SIlist{6.8;6.4;6.2;6.01}{\centi\meter}, using the apparatus shown in Fig.~\ref{fig-terrain-experiment}(A). The last diameter is an estimate for the smallest hole a \SI{6.0}{\centi\meter} payload could pass, considering the robot material.
 
\begin{equation}
   D_{\mathrm{hole}} = D_{\mathrm{payload}} + \frac{4tD_{\mathrm{robot}}}{D_{\mathrm{payload}}}.
    \label{eq:aspect_ratio}
\end{equation}
These apertures provide clearances from \SI{0.01}{\centi\meter} to \SI{0.8}{\centi\meter}.The robot body was pre-inserted through the hole to attribute any failure solely to object transport limitations.

For each object-aperture pairing, ten trials were performed with randomized initial orientations. Preliminary observations showed that object orientation critically affects success when clearance is minimal, as slight misalignments can induce jamming. However, for some objects, the ground plane and the tail’s positive pressure seal aligned them more reliably than expected, consistently orienting them before entering the hole. The success rates across ten trials are summarized in Fig.~\ref{fig-terrain-experiment}(B). All objects, except the cylinder, achieved a \SI{100}{\percent} success rate for a \SI{6.4}{\centi\meter} aperture. The cylinder failed because its most stable alignment was on its side (Fig.~\ref{fig-terrain-experiment}(C)), preventing it from passing through the hole. Increasing the cylinder’s height promoted axial alignment, suggesting tailoring object geometry for self-alignment. Both the cone and the sphere reached \SI{100}{\percent} success at the \SI{6.2}{\centi\meter} aperture, though these trials produced noticeable wear on the robot membrane.  At the smallest clearance (\SI{6.01}{\centi\meter}), some success was achieved but at the cost of significant membrane wear at the hole edge.


\subsubsection{Gap Crossing}
\label{subsubsec:gap-crossing-experiment}

Our final experiment tested the robot’s ability to bridge an unsupported horizontal gap while carrying a payload. First, the robot grew outward from one platform in a cantilevered configuration. Then, it spanned the gap before transporting the payload in a two-end supported configuration. The procedure tested the model in Eqn.~(\ref{eq:sum-moments}). In each trial, the robot started on a fixed platform and grew outward until the payload reached the end or the robot buckled (Fig. \ref{fig-terrain-experiment}(D)). The \kj{\(m_{payload}\)} varied from \(\SIrange[]{0}{200}{\gram}\) and the internal inflation pressure \(\kj{P_R}\) was artificially increased to determine the maximum length before collapse. The growth, including vertical tip deflection and length, were video recorded.

The data (Fig. \ref{fig-terrain-experiment}(E)) shows clear effects of payload and pressure on failure length. A critical length exists for each pressure, beyond which the structure can not support its weight. This limit decreases with payload and increases with pressure. These observations supplement moment-balance relations in Eqns.~\eqref{eq:critical-collapse-moment}–\eqref{eq:sum-moments}. The models over-predict collapse lengths, possibly due to the conditions leading to easier buckling when the robot left the table surface.

A follow-up test investigated the robot’s ability to move payloads over larger gaps. The robot grew across a gap to a surface \(\SI{15}{\centi\metre}\) below the starting platform.  After contact with the opposite edge, a \(\SI{500}{\gram}\) mass was carried inside the tail, and the robot attempted to grow further. This payload is heavier than the \(\SI{200}{\gram}\) limit used in the cantilevered trials. However, in Fig.~\ref{fig-terrain-experiment} (F), the robot successfully carried the \(\SI{500}{\gram}\) mass across a \(\SI{115}{\centi\metre}\) gap.
Improved performance is attributed to different loading conditions. During the three-point bending scenario, the lower membrane surface remained in tension, supporting a larger load than the cantilevered configuration, which placed the lower surface in compression. These results suggest that controlling object placement within the robot, possibly through intentional slipping within the tail, can extend the terrains that can be crossed.


\section{Discussion}\label{sec:discussion}
The experimental results in Section~\ref{sec:experimental} support the theoretical analysis in Section~\ref{sec:modeling} and define the operating envelope of the soft everting robot. In the delivery of object trials, the soft everting robot was able to both accurately and precisely deliver a payload to a desired location. The uncluttered environment was most evident, but even in a cluttered environment, complex phenomena like payload slippage, local buckling, and variable capstan friction resulted in moderately successful outcomes, even without active control. Introducing more active payload delivery strategies, such as modulation of internal flow and strategic pressurization, stands to improve results even further. These delivery trials demonstrate the beneficial nature of the presented base stations' in-situ object loading ability. Inclined‐surface tests with servos further confirmed the linear dependence of the required $\kj{P_R}$ on $m_{\text{payload}} g \sin\theta$, validating the analysis in Eqn.~\eqref{eq:inclined_growth} and Figure~\ref{fig-Incline-Analysis}. A small vertical intercept in the pressure–force fit represents an orientation‐independent loss $F_{\mathrm{loss}}$ which matches previous experimental results. Vertical climbs exposed bend‐induced jamming, validating Figure~\ref{fig-Bend-Analysis} and consistent with the results of testing delivery around bends. Nevertheless, the robot lifted a \SI{1.5}{\kilogram} payload vertically, demonstrating the efficacy soft everting robots for payload delivery.

Securing experiments (Section~\ref{subsec: secure-experiment}) upheld the extended gripper model. The measured linear force–pressure curves matched Eqn.~\eqref{eq:K_model} once the membrane‐wrap term \(\tfrac12 A_c\cos\phi\) was included. Shape‐series tests showed that, with an appropriate \kj{\(A_w\)} estimate, the same model predicts holding forces for a range of shapes. Hence gripping performance for novel geometries can be predicted quantitatively.

Finally, terrain studies show that many beneficial features of movement by growth are preserved when delivering payloads in this manner, and the results suggest some mitigation strategies for observed effects. Hole‐restriction trials revealed that near‐zero clearance yields intermittent jams, but utilizing Eqn.~\eqref{eq:aspect_ratio} elevates success to nearly \(100\%\).  Ground contact and tail tension can align objects if shaped well, indicating that deliberate orientation control could further reduce jamming.
Gap‐crossing tests validated the critical span length. Higher pressure extended reach but with diminishing benefit, bounded by material strength. The results furnish a pragmatic design rule: for a specified gap and payload, trajectory adjustment or supplemental supports can better guarantee success.  

Overall, the results indicate that payload delivery using soft everting robots is a promising method to deliver large or heavy payloads in addition to sensors and actuators. There were some limitations found. The primary limitation to carrying payloads was material friction, which the servo’s limitations prevented it from overcoming, causing significant material wear. Additionally, causing a payload to emerge at a desired point, shown in Section~\ref{subsubsection:Delivery}, required precise timing and properly orientate. Achieving practical repositioning of passive payloads requires new planning strategies or active payloads.


\section{CONCLUSION}\label{sec:conclusion}

This work delivers an integrated framework that links theory with practice for payload delivery using a soft everting robot. Analytical models establish conditions for successful growth while carrying a load, predict the gripping force that restrains an object within the tail, and outline the pressure and span limits that govern passive gap crossing. A comprehensive set of trials confirms those predictions and shows that the robot can transport objects up to $\SI{1500}{\gram}$ through slopes, tight apertures, and unsupported gaps with only minor path adjustments. The agreement between the model and the data demonstrates that the key mechanical factors are now understood well enough to guide mission planning and payload packaging. Future studies will extend the approach to dynamic maneuvers, incorporate variable‑stiffness skins to suppress sag, and investigate closed‑loop strategies that couple onboard sensing with autonomous routing, broadening the range of tasks accessible to soft everting robots in real environments.

\bibliographystyle{unsrt}
\bibliography{References}

@article{blumenschein2020design,
  title={Design, modeling, control, and application of everting vine robots},
  author={Blumenschein, Laura H and Coad, Margaret M and Haggerty, David A and Okamura, Allison M and Hawkes, Elliot W},
  journal={Frontiers in Robotics and AI},
  volume={7},
  pages={548266},
  year={2020},
  publisher={Frontiers Media SA}
}

@inproceedings{haggerty2019characterizing,
  title={Characterizing environmental interactions for soft growing robots},
  author={Haggerty, David A and Naclerio, Nicholas D and Hawkes, Elliot W},
  booktitle={2019 IEEE/RSJ International Conference on Intelligent Robots and Systems (IROS)},
  pages={3335--3342},
  year={2019},
  organization={IEEE}
}

@ARTICLE{origamiRyu,
  author={Kim, Ji-hun and Jang, Jaehyung and Lee, Sang-min and Jeong, Sang-Goo and Kim, Yong-Jae and Ryu, Jee-Hwan},
  journal={IEEE Robotics and Automation Letters}, 
  title={Origami-inspired {N}ew {M}aterial {F}eeding {M}echanism for {S}oft {G}rowing {R}obots to {K}eep the {C}amera {S}tay at the {T}ip by {S}ecuring its {P}ath}, 
  year={2021},
  volume={6},
  number={3},
  pages={4592-4599},
  doi={10.1109/LRA.2021.3068936}}

@ARTICLE{vineRobotsReview,
  author={Coad, Margaret M. and Blumenschein, Laura H. and Cutler, Sadie and Zepeda, Javier A. Reyna and Naclerio, Nicholas D. and El-Hussieny, Haitham and Mehmood, Usman and Ryu, Jee-Hwan and Hawkes, Elliot W. and Okamura, Allison M.},
  journal={IEEE Robotics \& Automation Magazine}, 
  title={Vine {R}obots}, 
  year={2020},
  volume={27},
  number={3},
  pages={120-132},
  doi={10.1109/MRA.2019.2947538}}

@INPROCEEDINGS{roboa,
  author={der Maur, Pascal Auf and Djambazi, Betim and Haberthür, Yves and Hörmann, Patricia and Kübler, Alexander and Lustenberger, Michael and Sigrist, Samuel and Vigen, Oda and Förster, Julian and Achermann, Florian and Hampp, Elias and Katzschmann, Robert K. and Siegwart, Roland},
  booktitle={IEEE International Conference on Soft Robotics}, 
  title={{RoBoa}: {C}onstruction and {E}valuation of a {S}teerable {V}ine {R}obot for {S}earch and {R}escue {A}pplications}, 
  year={2021},
  volume={},
  number={},
  pages={15-20},
  doi={10.1109/RoboSoft51838.2021.9479192}}

@article{hawkes2017soft,
author = {Elliot W. Hawkes  and Laura H. Blumenschein  and Joseph D. Greer  and Allison M. Okamura },
title = {A soft robot that navigates its environment through growth},
journal = {Science Robotics},
volume = {2},
number = {8},
pages = {eaan3028},
year = {2017},
doi = {10.1126/scirobotics.aan3028},
URL = {https://www.science.org/doi/abs/10.1126/scirobotics.aan3028},
}

@ARTICLE{coad2020retraction,
  author={Coad, Margaret M. and Thomasson, Rachel P. and Blumenschein, Laura H. and Usevitch, Nathan S. and Hawkes, Elliot W. and Okamura, Allison M.},
  journal={IEEE Robotics and Automation Letters}, 
  title={Retraction of {S}oft {G}rowing {R}obots {W}ithout {B}uckling}, 
  year={2020},
  volume={5},
  number={2},
  pages={2115-2122},
  doi={10.1109/LRA.2020.2970629}}

@article{kim2023soft,
author = {Kim, Nam Gyun and Ryu, Jee-Hwan},
title = {A {S}oft {G}rowing {R}obot {U}sing {H}yperelastic {M}aterial},
journal = {Advanced Intelligent Systems},
volume = {5},
number = {2},
pages = {2200264},
doi = {https://doi.org/10.1002/aisy.202200264},
}

@article{SoftBurrowingRobot,
author = {Nicholas D. Naclerio  and Andras Karsai  and Mason Murray-Cooper  and Yasemin Ozkan-Aydin  and Enes Aydin  and Daniel I. Goldman  and Elliot W. Hawkes },
title = {Controlling subterranean forces enables a fast, steerable, burrowing soft robot},
journal = {Science Robotics},
volume = {6},
number = {55},
pages = {eabe2922},
year = {2021},
doi = {10.1126/scirobotics.abe2922},
}

@INPROCEEDINGS{coralReefs,
  author={Luong, Jamie and Glick, Paul and Ong, Aaron and deVries, Maya S. and Sandin, Stuart and Hawkes, Elliot W. and Tolley, Michael T.},
  booktitle={IEEE International Conference on Soft Robotics}, 
  title={Eversion and {R}etraction of a {S}oft {R}obot {T}owards the {E}xploration of {C}oral {R}eefs}, 
  year={2019},
  volume={},
  number={},
  pages={801-807},
  doi={10.1109/ROBOSOFT.2019.8722730}}

@INPROCEEDINGS{tipMount,
  author={Jeong, Sang-Goo and Coad, Margaret M. and Blumenschein, Laura H. and Luo, Ming and Mehmood, Usman and Kim, Ji Hun and Okamura, Allison M. and Ryu, Jee-Hwan},
  booktitle={IEEE/RSJ International Conference on Intelligent Robots and Systems}, 
  title={A {T}ip {M}ount for {T}ransporting {S}ensors and {T}ools using {S}oft {G}rowing {R}obots}, 
  year={2020},
  volume={},
  number={},
  pages={8781-8788},
  doi={10.1109/IROS45743.2020.9340950}}

@article{steerableCamera,
author = {Greer, Joseph D. and Morimoto, Tania K. and Okamura, Allison M. and Hawkes, Elliot W.},
title = {A Soft, Steerable Continuum Robot That Grows via Tip Extension},
journal = {Soft Robotics},
volume = {6},
number = {1},
pages = {95-108},
year = {2019},
doi = {10.1089/soro.2018.0034},
}

@ARTICLE{vineRobotsReview2,
AUTHOR={Blumenschein, Laura H.  and Coad, Margaret M.  and Haggerty, David A.  and Okamura, Allison M.  and Hawkes, Elliot W. },
TITLE={Design, {M}odeling, {C}ontrol, and {A}pplication of {E}verting {V}ine {R}obots},
JOURNAL={Frontiers in Robotics and AI},
VOLUME={7},
YEAR={2020},
DOI={10.3389/frobt.2020.548266},

}

@INPROCEEDINGS{softCameraHolder,
  author={Heap, William E. and Naclerio, Nicholas D. and Coad, Margaret M. and Jeong, Sang-Goo and Hawkes, Elliot W.},
  booktitle={IEEE/RSJ International Conference on Intelligent Robots and Systems}, 
  title={Soft {R}etraction {D}evice and {I}nternal {C}amera {M}ount for {E}verting {V}ine {R}obots}, 
  year={2021},
  volume={},
  number={},
  pages={4982-4988},
  doi={10.1109/IROS51168.2021.9636697}}

@article{rus2015design,
  title={Design, fabrication and control of soft robots},
  author={Rus, Daniela and Tolley, Michael T},
  journal={Nature},
  volume={521},
  number={7553},
  pages={467--475},
  year={2015},
  publisher={Nature Publishing Group UK London}
}

@Inbook{della2020soft,
author="Della Santina, Cosimo
and Catalano, Manuel G.
and Bicchi, Antonio",
editor="Ang, Marcelo H.
and Khatib, Oussama
and Siciliano, Bruno",
title="Soft Robots",
bookTitle="Encyclopedia of Robotics",
year="2020",
publisher="Springer Berlin Heidelberg",
address="Berlin, Heidelberg",
pages="1--15",
isbn="978-3-642-41610-1",
doi="10.1007/978-3-642-41610-1_146-2",
url="https://doi.org/10.1007/978-3-642-41610-1_146-2"
}

@INPROCEEDINGS{suulker2023soft,
  author={Suulker, Cem and Skach, Sophie and Kaleel, Danyaal and Abrar, Taqi and Murtaza, Zain and Suulker, Dilara and Althoefer, Kaspar},
  booktitle={IEEE/RSJ International Conference on Intelligent Robots and Systems}, 
  title={Soft {C}ap for {V}ine {R}obots}, 
  year={2023},
  volume={},
  number={},
  pages={6462-6468},
  doi={10.1109/IROS55552.2023.10341377}}

@INPROCEEDINGS { OseleICRA2024,
    author={Osele, O. Godson and Barhydt, Kentaro and Cerone, Nicholas and Okamura, Allison M. and Harry Asada, H.},
    TITLE = { Tip-{C}lutching {W}inch for {H}igh {T}ensile {F}orce {A}pplication with {S}oft {G}rowing {R}obots },
    BOOKTITLE = { IEEE International Conference on Robotics and Automation },
    YEAR = { 2024 },
}

@article{laschi2016soft,
author = {Cecilia Laschi  and Barbara Mazzolai  and Matteo Cianchetti },
title = {Soft robotics: {T}echnologies and systems pushing the boundaries of robot abilities},
journal = {Science Robotics},
volume = {1},
number = {1},
pages = {eaah3690},
year = {2016},
doi = {10.1126/scirobotics.aah3690},
URL = {https://www.science.org/doi/abs/10.1126/scirobotics.aah3690},
}

@article{sui2022bioinspired,
  title={A bioinspired soft swallowing gripper for universal adaptable grasping},
  author={Sui, Dongbao and Zhu, Yanhe and Zhao, Sikai and Wang, Tianshuo and Agrawal, Sunil K and Zhang, He and Zhao, Jie},
  journal={Soft Robotics},
  volume={9},
  number={1},
  pages={36--56},
  year={2022},
  publisher={Mary Ann Liebert, Inc., publishers 140 Huguenot Street, 3rd Floor New~…}
}

@article{root2021bio,
  title={Bio-inspired design of soft mechanisms using a toroidal hydrostat},
  author={Root, Samuel E and Preston, Daniel J and Feifke, Gideon O and Wallace, Hunter and Alcoran, Renz Marion and Nemitz, Markus P and Tracz, Jovanna A and Whitesides, George M},
  journal={Cell Reports Physical Science},
  volume={2},
  number={9},
  year={2021},
  publisher={Elsevier}
}

@article{li2025variable,
  title={A Variable Stiffness Bioinspired Swallowing Gripper Based on Particle Jamming},
  author={Li, Mingge and Huang, Xiaoming and Liu, Quan and Yin, Zhongjun},
  journal={Soft Robotics},
  volume={12},
  number={1},
  pages={56--67},
  year={2025},
  publisher={Mary Ann Liebert, Inc., publishers 140 Huguenot Street, 3rd Floor New~…}
}

@article{li2020bioinspired,
  title={A bioinspired soft swallowing robot based on compliant guiding structure},
  author={Li, Haili and Yao, Jiantao and Liu, Chunye and Zhou, Pan and Xu, Yundou and Zhao, Yongsheng},
  journal={Soft Robotics},
  volume={7},
  number={4},
  pages={491--499},
  year={2020},
  publisher={Mary Ann Liebert, Inc., publishers 140 Huguenot Street, 3rd Floor New~…}
}

@article{comer1963deflections,
  title={Deflections of an inflated circular-cylindrical cantilever beam},
  author={Comer, RL and Levy, Samuel},
  journal={AIAA journal},
  volume={1},
  number={7},
  pages={1652--1655},
  year={1963}
}

@inproceedings{mcfarland2023collapse,
  title={Collapse of straight soft growing inflated beam robots under their own weight},
  author={McFarland, Ciera and Coad, Margaret M},
  booktitle={IEEE International Conference on Soft Robotics},
  pages={1--8},
  year={2023},
}

@book{leonard1960structural,
  title={Structural considerations of inflatable reentry vehicles},
  author={Leonard, Robert W},
  year={1960},
  publisher={National Aeronautics and Space Administration}
}

@ARTICLE{Seo2024WorkingChannel,
  author={Seo, Dongoh and Kim, Nam Gyun and Ryu, Jee-Hwan},
  journal={IEEE Robotics and Automation Letters}, 
  title={Inflatable-Structure-Based Working-Channel Securing Mechanism for Soft Growing Robots}, 
  year={2024},
  volume={9},
  number={9},
  pages={7755-7762},
  doi={10.1109/LRA.2024.3426322}}

@INPROCEEDINGS{Boateng2024SSRR,
  author={Boateng, Sidney Nimako and Singh, Shashwat and Ugur, Mustafa and Wang, Sicheng and Kramer, Max and Osei, Isaac and Sobek, Olivia and Martinez, Melisa Orta and Temel, F. Zeynep and Blumenschein, Laura H.},
  booktitle={IEEE International Symposium on Safety Security Rescue Robotics}, 
  title={Heterogenous Collaboration: A new approach for search and rescue operations}, 
  year={2024},
  volume={},
  number={},
  pages={142-147},
  doi={10.1109/SSRR62954.2024.10770037}}

@article{giri2025inchigrab,
  title={InchIGRAB: An inchworm-inspired guided retraction and bending device for vine robots during colonoscopy},
  author={Giri, Ayush and Girerd, C{\'e}dric and Cervera-Torralba, Jacobo and Tolley, Michael T and Morimoto, Tania K},
  journal={IEEE/ASME Transactions on Mechatronics},
  year={2025},
}

@book{blumenschein2019design,
  title={Design and Modeling of Soft Growing Robots},
  author={Blumenschein, Laura H},
  year={2019},
  publisher={Stanford University}
}

@article{heap2024large,
  title={Large-scale vine robots for industrial inspection: Developing a new framework to overcome limitations with existing inspection methods},
  author={Heap, William E and Man, Steven and Bassari, Vedad and Nguyen, Steven and Yao, Elvy B and Tripathi, Neel A and Naclerio, Nicholas D and Hawkes, Elliot W},
  journal={IEEE Robotics \& Automation Magazine},
  year={2024},
  publisher={IEEE}
}

@article{McFarland2025gap,
        title={Modeling Collapse of Steered Vine Robots Under Their Own Weight}, 
      author={Ciera McFarland and Margaret McGuinness},
      year={2025},
      eprint={2510.25727},
      archivePrefix={arXiv},
      primaryClass={cs.RO},
      url={https://arxiv.org/abs/2510.25727}, 
}

@inproceedings{godaba2019payload,
  title={Payload capabilities and operational limits of eversion robots},
  author={Godaba, Hareesh and Putzu, Fabrizio and Abrar, Taqi and Konstantinova, Jelizaveta and Althoefer, Kaspar},
  booktitle={Annual Conference Towards Autonomous Robotic Systems},
  pages={383--394},
  year={2019},
  organization={Springer}
}

@inproceedings{park2025soft,
  title={Soft Everting Prosthetic Hand and Comparison with Existing Body-Powered Terminal Devices},
  author={Park, Gayoung and Sch{\"a}affer, Katalin and Coad, Margaret M},
  booktitle={2025 IEEE 8th International Conference on Soft Robotics (RoboSoft)},
  pages={1--7},
  year={2025},
  organization={IEEE}
}

\end{document}